\documentclass{article}

\usepackage[nonatbib, final]{nips_2017}

\usepackage[utf8]{inputenc} 
\usepackage[T1]{fontenc}    
\usepackage{hyperref}       
\usepackage{url}            
\usepackage{booktabs}       
\usepackage{amsfonts}       
\usepackage{nicefrac}       
\usepackage{microtype}      
\usepackage{todonotes}      
\usepackage{algpseudocode}
\usepackage{algorithm}  
\usepackage{cite}
\usepackage{amsmath}
\usepackage{bm}
\usepackage{bbm}
\usepackage{graphicx}
\usepackage{pgfplots}
\usepgfplotslibrary{groupplots}
\usepackage{siunitx}

\DeclareMathOperator*{\argmax}{arg\,max}
\newcommand{\rpm}{\raisebox{.2ex}{$\scriptstyle\pm$}}

\title{Active Robotic Mapping through Deep Reinforcement Learning}

\author{
  Shane Barratt \\
  Department of Electrical Engineering\\
  Stanford University \\
  Stanford, CA 94305 \\
  \texttt{sbarratt@stanford.edu} \\
}

\begin{document}

\maketitle

\begin{abstract}
We propose an approach to learning agents for active robotic mapping, where the goal is to map the environment as quickly as possible. The agent learns to map efficiently in simulated environments by receiving rewards corresponding to how fast it constructs an accurate map. In contrast to prior work, this approach learns an exploration policy based on a user-specified prior over environment configurations and sensor model, allowing it to specialize to the specifications. We evaluate the approach through a simulated Disaster Mapping scenario and find that it achieves performance slightly better than a near-optimal myopic exploration scheme, suggesting that it could be useful in more complicated problem scenarios.
\end{abstract}

\section{Introduction}
For most of the paper, we will use the following terminology borrowed from \cite{thrun2005probabilistic}:
\begin{itemize}
\itemsep0em
\item robot: an autonomous agent in an environment
\item pose $(x_t)$: position and orientation of the robot in that environment
\item observations $(z_t)$: readings from on-board sensors
\item controls $(u_t)$: control inputs to actuators
\item policy ($\pi(u_t|x_{1:t-1}, z_{1:t})$): method for choosing controls based on past poses and observations
\end{itemize}

Autonomous robots face two fundamental problems: \emph{mapping} and \emph{localization}. The goal of mapping is to obtain a spatial model of the robot's environment using poses and observations. Simultaneously, the goal of localization is to estimate the robot's pose using observations and a map. These problems are inherently linked, as mapping requires poses and localization requires maps. Addressing both is commonly referred to as the simultaneous localization and mapping (SLAM) problem \cite{leonard1991mobile}. In this work, we focus on the problem of efficient mapping with known poses, which assumes that the localization problem has already been solved for us. We view the integration of localization into our approach as future work.

There are three main hurdles we must overcome when designing robotic mapping systems. First, we require a representation of the map as well as our belief over possible maps. Second, we need a way to update our belief over maps given observations. Third, we need to choose controls to reduce our uncertainty in our belief as quickly as possible.

One of the most widely used (and elegant) ways to represent the map probabilistically and incorporate observations sequentially is via occupancy grids \cite{elfes1989using}. Occupancy grids exhibit several desirable properties: they are adaptable for many applications, they compute an exact posterior distribution over maps (under some independence assumptions), they are strongly convergent, they can deal with any type of sensor noise, and they can handle raw sensor data. However, occupancy grid maps lack a principled method for our third hurdle: how to choose control inputs. The original proposal assumes that control inputs are specified by a third-party and not chosen by the robot. This leads to an issue when applying occupancy grids in practice; how should the actions be chosen?

We pose the problem of designing agents that actively map the environment as a Markov Decision Process (MDP), where the state of this MDP is the robot's pose and its belief of the environment. From this state, the robot should be able to identify areas in the map that require exploration and to move to them and map them, based on its knowledge of how its sensors work. The robot's belief is a high-dimensional matrix of probability values. Accordingly, we turn to deep reinforcement learning techniques to find policies. To train such policies, we use a prior over maps the robot might encounter and repeatedly train the robot to explore these maps in simulation. We evaluate our algorithm on a simulated Disaster Mapping environment, and find that it is able to achieve performance on-par with a near-optimal greedy strategy. This suggests that the algorithm will be able to scale to more complex environments and sensors, which we view as future work.

\section{Related Work}

The two most popular methods for addressing the third hurdle in occupancy grids are frontier-based exploration \cite{yamauchi1997frontier}, where the robot actively seeks to visit new poses, and information-based exploration \cite{bourgault2002information}, where the robot myopically chooses control inputs to maximize information gain over one step.

\subsection{Frontier-Based Exploration}

In frontier-based exploration, the robot maintains a set of \emph{frontiers}, which are defined as regions on the boundary between open and unexplored space. The robot then navigates to the closest frontier by performing a depth-first-search (DFS) on the maximum likelihood map. If, after a predetermined number of steps, the robot fails to reach the frontier region it repeats the process over again. There is also a natural multi-robot extension\cite{yamauchi1998frontier}.

\subsection{Information-Based Adaptive Robotic Exploration}
\label{sec:igain}

In information-based exploration, the robot moves in the direction that maximizes the expected information gain

$$
\argmax_{u_t} H(b_t) - \mathbb{E}_{x_{t+1}, z_{t+1}}[H(b_{t+1})|u_t, x_t],
$$

where

$$\mathbb{E}_{x_{t+1}, z_{t+1}}[H(b_{t+1})|u_t, x_t] = \sum_{x_{t+1}} \sum_{z_{t+1}} P(x_{t+1}|x_t, u_t) P(z_{t+1}|x_{t+1}) H(b_{t+1}|x_{t+1}, z_{t+1}).$$

Note that this summation can become extremely hard to compute, as the summation becomes an integral when the sensor outputs continuous values and the summation can be exponential with the number of possible sensor outputs. Also, this is a myopic exploration strategy; it only looks one step into the future to decide its next action. However, in discrete environments with simple sensors it can be a near-optimal exploration scheme. This is the baseline that we compare our approach to.

\section{Preliminaries}

In this section, we introduce Occupancy Grids, Markov Decision Processes and the Advantage Actor-Critic Algorithm. The section also serves as an introduction of the mathematical notation that will be used throughout the paper.

\subsection{Occupancy Grids}

The basic idea behind occupancy grid maps is to represent the map as a two-dimensional grid of binary random variables that represent whether or not the location is \emph{occupied} or \emph{not occupied}.

More concretely, we posit that there is an underlying grid map $m \in M = \{0, 1\}^{N \times N}$ that is a-priori unknown to the robot. We wish to calculate our belief over maps $M$ at time $t$ given all previous poses and observations leading up to that time step $b_t(m) = p(m|x_{1:t}, z_{1:t})$. Reasoning about all possible maps quickly becomes intractable, as there are $2^{N^2}$ possible maps. To simplify the problem, we assume the individual map random variables, indexed as $m_i$, are independent given the poses and measurements, or

$$b_t(m)= \prod_{i} p(m_i|x_{1:t}, z_{1:t}) = \prod_{i} b_t(m_i).$$

We work with the log-odds posterior $l_{t, i} = \log \frac{b_t(m_i=1)}{1 - b_t(m_i=1)}$ for simplicity and can recover our posterior probabilities using $b_t(m_i=1) = 1 - \frac{1}{1 + \exp\{l_{t, i}\}}$. We can update our posterior given a sensor reading via an inverse sensor model $p(m_i | x_t, z_t)$ using a recursive Bayes filter \cite{thrun2005probabilistic}:

\begin{equation}l_{t, i} = l_{t-1, i} + \log \frac{p(m_i=1 | x_t, z_t)}{1-p(m_i=1 | x_t, z_t)} - \log \frac{p(m_i = 1)}{p(m_i = 0)}.
\label{recbayesfilter}
\end{equation}

Later, we will use the information-theoretic entropy of the belief state to quantify our uncertainty, which factorizes over the individual map random variables because they are assumed to be independent,

$$H(b_t(m)) = \sum_i H(b_t(m_i)) = \sum_i b_t(m_i = 1) \log b_t(m_i = 1) + b_t(m_i = 0) \log b_t(m_i = 0).$$

\subsection{Markov Decision Processes}
\label{sec:mdp}

Markov decision processes (MDP) are a mathematical framework for decision making where the outcomes are random and potentially affected by the decision maker \cite{sutton1998reinforcement}. An MDP is formally defined as a tuple $<S, A, T, R, \gamma>$. $S$ is a set of states the decision maker might be in. $A$ is the set of actions the decision maker can take. $T(s'|s, a)$ is the distribution over next states given the decision maker took action $a$ in state $s$. $R(s, a)$ is the reward the agent receives for taking action $a$ in state $s$. $\gamma$ is the discount factor, weighing short-term rewards versus long-term rewards.

\subsection{Advantage Actor-Critic}
\label{sec:a2c}

The Advantage Actor-Critic (A2C) algorithm is the single-threaded version of \cite{mnih2016asynchronous}, and has enjoyed lots of success in playing classic ATARI games from raw pixel observations. It prescribes a method for learning a differentiable stochastic policy $\pi(a|s; \theta')$ and value function $V(s; \theta_v')$ through interactions with an environment. More concretely, after $n$ interactions with the environment, it performs a gradient step on the following loss functions for each interaction:

$$L(\theta) = -\log \pi(a_t|s_t; \theta')(R-V(s_t;\theta_v')) - \lambda H(\pi(a_t|s_t; \theta')),$$

$$L(\theta_v') = (R-V(s_t;\theta_v'))^2,$$

where $R$ is the bootstrapped $n-$step reward. We use the A2C algorithm to train our agents.

\section{Approach}

We separate the problem of updating the posterior of the belief from selecting actions. At each time step $t$, the robot receives an observation $z_t$ and its pose $x_t$. It updates its posterior $b_t$ using Equation~\ref{recbayesfilter}. In other words, the agent only has control over $u_t$ and not how $b_t$ is updated. At this point, the robot is faced with the decision of what control input $u_t$ to select given $b_t$ and $x_t$. Instead of choosing the action myopically or planning, we resort to flexible model-free reinforcement learning to train our mapping agent to act directly based on its belief state.

\begin{figure}[h]
    \centering
    \label{fig:dynamics}
    \caption{A full transition of our mapping environment.}
    \medskip
    \includegraphics[scale=.4]{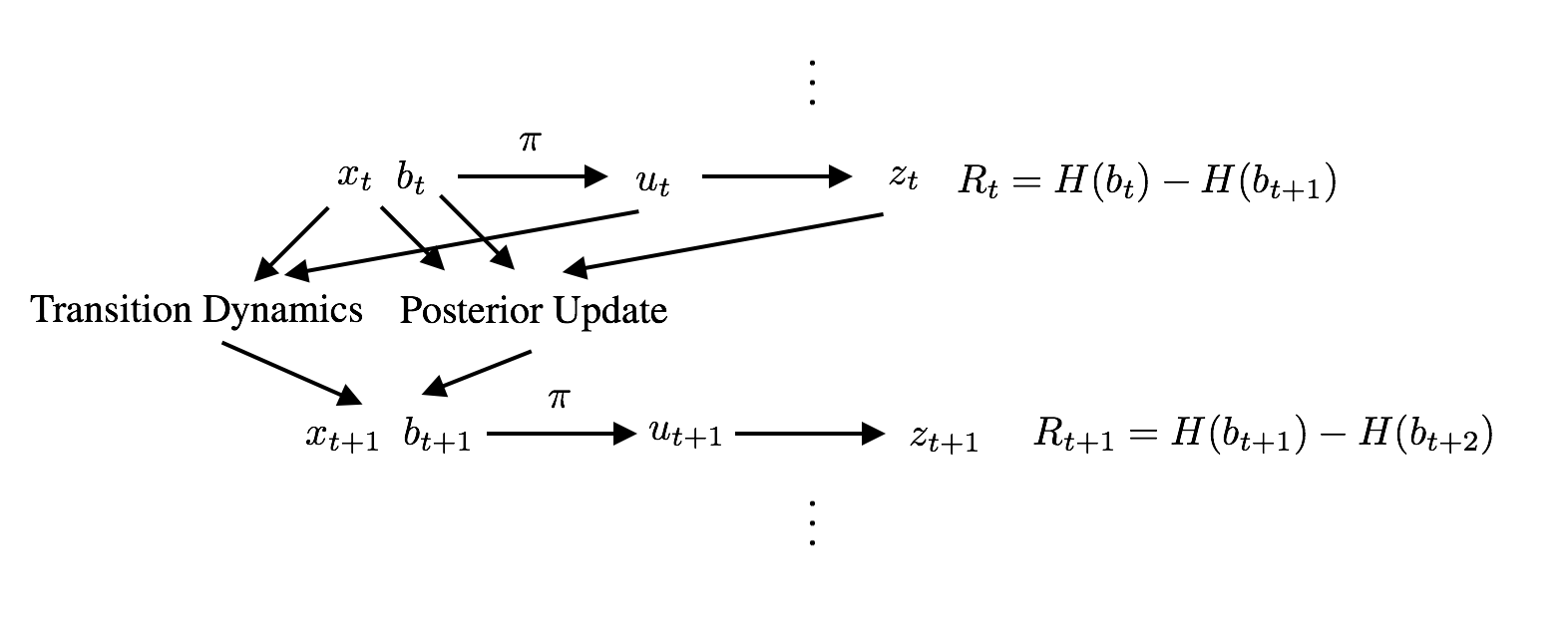}
\end{figure}

We convert our notation to the MDP notation in Section~\ref{sec:mdp}. We let the state be defined as $s_t=[b_t, x_t]$ and the action as $a_t=u_t$. The state evolves according to Equation~\ref{recbayesfilter} and the underlying map. In our RL formulation, the robot seeks a policy $\pi(a|s)$ that reduces uncertainty as quickly as possible, or equivalently one that maximizes the following discounted expected cumulative reward

$$\mathbb{E}_{\pi, m} \big [ \sum_{t=0}^{T-1} \gamma^t R_t \big ],$$

where $R_t=H(b_t) - H(b_{t+1})$ is the reduction in the entropy of the robot's belief state at time step $t$ and the expectation is taken over maps, the policy and the imposed MDP dynamics. The dynamics of the environment are presented in Figure~\ref{fig:dynamics}.

We can then train a policy to maximize the expected reward in simulation using a prior over initial maps and the A2C algorithm from Section~\ref{sec:a2c}. Our prior over initial maps can be thought of as a representative distribution over the environments our robot might encounter. We could learn this distribution by training a generative model over a database of previously acquired maps, or hand-design it. The full learning procedure is summarized in Algorithm~\ref{alg:l2map}.

This learning procedure can accommodate a variety of posterior updates, priors over maps and pose/control spaces, as long as the architecture of the actor-critic is chosen correctly.

\begin{algorithm}[b!]
  \caption{Learning to Map}
  \label{alg:l2map}
  \begin{algorithmic}[1]
    \Require $p(m)$, prior distribution over maps. $p(x)$, prior distribution over initial poses. $p(m|x, z)$, inverse sensor model. $\pi$ and $V$, differentiable actor and critic.
    \For{$N$ episodes}
    \State Sample $m \sim p(m)$
    \State Simulate episode updating $\pi$ and $V$ using A2C algorithm 
    \EndFor
\end{algorithmic}
\end{algorithm}

\subsection{Why is the posterior update part of the environment?}

This formulation treats the posterior update as part of the environment, which seems counter-intuitive as the robot normally updates its posterior itself. However, we assume we already have a sensible posterior update rule and we seek a complementary policy that can reduce uncertainty as quickly as possible.

\section{Application: Disaster Mapping}

\begin{figure}[H]
    \centering
    \label{fig:env}
    \caption{The Disaster Mapping Environment. Left, the fully observable environment with the quad-copter (pink) and buildings (gray). Right, the belief state of the robot with the quad-copter (pink) where occupancy probabilities are represented as a gradient from black (0) to green (1).}
    \medskip
    \includegraphics[scale=.25]{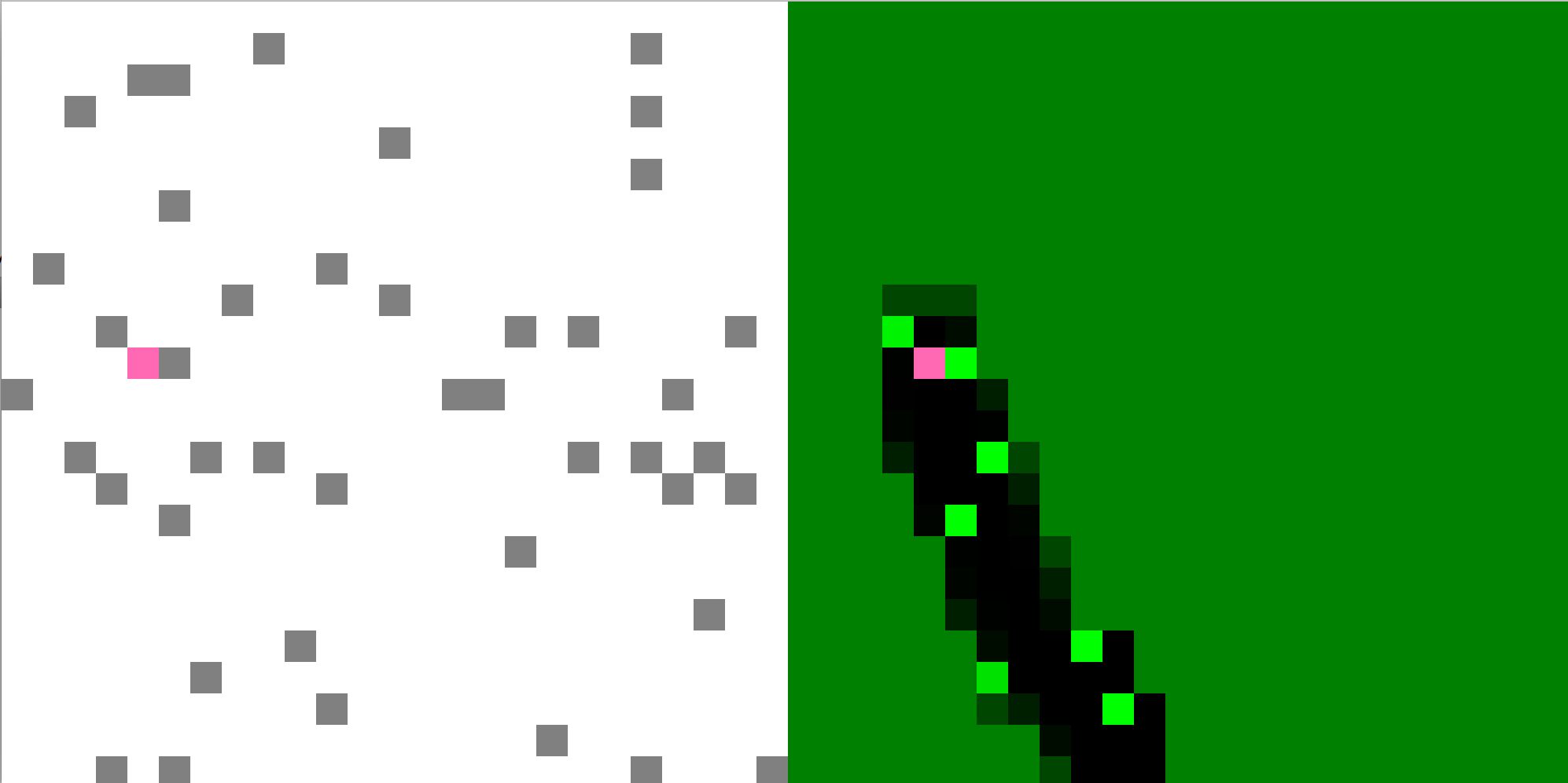}
\end{figure}

We will use the following motivating scenario to set up an environment to test our approach. A large earthquake has occurred on the San Andreas Fault, decimating many of the buildings in San Francisco. We would like to find out which buildings are still standing, but the entire city is covered in smoke so we cannot use imaging from planes or satellites. Luckily, we have a quad-copter that is able to fly through the smoke and get close to the buildings. This is a robotic mapping problem. We represent the map of buildings as a $25 \times 25$ occupancy grid. We discretize the location of the drone so that it is in a single position in the grid and assume it is flying at a fixed altitude. The drone can sense whether or not there is a building in the adjacent positions, but has a noisy sensor that is wrong 20\% of the time. The drone cannot can move to adjacent positions that do not contain buildings. We have an independent Bernoulli($.1$) prior over maps. Figure~\ref{fig:env} is a visual representation of the environment.

\subsection{Architecture Design}

To use A2C, we need a differentiable policy $\pi(a_t|b_t, x_t; \theta)$ and value function $V(b_t, x_t; \theta')$. For the case of disaster mapping, our belief state is a two-dimensional grid of occupancy probability values $b_t(m_i)$ and our pose $x_t$ is a position in that grid. In order to take advantage of the spatial representation of the belief and pose, we form a $N \times N$ matrix $B_t$ where each entry has its corresponding belief value $b_t(m_i)$. If our robot is at a pose $x_t=(i, j)$, we form a \emph{centered} version of the belief: a $(2N-1) \times (2N-1)$ matrix $C_t=B_t[i-(N-1):i+(N-1), j-(N-1):j+(N-1)]$. We pad $B$ with 1's before forming $C_t$, because we are sure that positions outside the map contains obstacles. We note that the index $(N, N)$ in $C_t$ always represents the robot's current pose and that all of $B$ is always present in $C_t$, thus making it a sufficient statistic of $[b_t, x_t]$.

This whole process can be made more concrete with an example. Suppose we have a $3 \times 3$ belief state $B_0$ that is $.5$ everywhere and $0$ at the current pose $x_0=(0, 0)$. Then $C_0$ is $5 \times 5$ and defined as

$$C_t =
\begin{bmatrix}
    1       & 1 & 1 & 1 & 1 \\
    1       & 1 & 1 & 1 & 1 \\
    1       & 1 & 0 & .5 & .5 \\
    1       & 1 & .5 & .5 & .5 \\
    1       & 1 & .5 & .5 & .5
\end{bmatrix} \ \
H(C_t) =
\begin{bmatrix}
    0       & 0 & 0 & 0 & 0 \\
    0       & 0 & 0 & 0 & 0 \\
    0       & 0 & 0 & .69 & .69 \\
    0       & 0 & .69 & .69 & .69 \\
    0       & 0 & .69 & .69 & .69
\end{bmatrix}
.$$

We propose to also add the point-wise entropy map of $C_t$ as a feature channel in our state, to allow our robot to guide itself to high entropy areas, resulting in the state $s_t=[C_t, H(C_t)]$.

We experiment with three different network architectures, of increasing complexity. They all result in a tensor $\in \mathbb{R}^{256}$ that is processed in parallel by a fully-connected (FC) layer followed by a soft-max for $\pi$ and a FC layer for $V$.

\paragraph{Multi-Layer Perceptron (MLP)} The input features $s_t$ are flattened and passed through a single FC layer to a hidden layer of size $256$ followed by a rectified linear unit (ReLU) nonlinearity.

\paragraph{Convolutional Neural Network (CNN-MLP)} The input features $s_t$ are processed by the following convolutional layers with ReLU nonlinearities:
\begin{enumerate}
\itemsep0em
\item[{(1)}] 32 $3 \times 3$ filters with stride 2
\item[{(2)}] 32 $4 \times 4$ filters with stride 2
\item[{(3)}] 32 $8 \times 8$ filters with stride 1
\item[{(4)}] A FC layer to a hidden layer of size $256$ followed by ReLU
\end{enumerate}

\paragraph{Residual Network (ResNet)} The input features $s_t$ are processed by a single convolutional layer then 6 residual blocks \cite{he2016deep} with ReLU nonlinearities. Each convolutional layer in the network has 64 $3 \times 3$ filters with stride $1$ and padding $1$. Finally, the flattened features are passed through a FC layer to a hidden layer of size $256$ followed by a ReLU.

All of the architectures are implemented in PyTorch \cite{PyTorch} and the code is available online\footnote{https://www.github.com/sbarratt/rl-mapping}. A video of a trained policy in action is also available online\footnote{https://www.youtube.com/watch?v=6m4U7mWNOzs}.

\subsection{Training Details}

We use a discount rate $\gamma=.99$ and episode length of $300$. The agent runs for $20$ steps until an A2C update. We run the learning algorithm ADAM for $10$k episodes, resulting in $100$k gradient updates to the actor-critic\cite{kingma2014adam}. We use a learning rate of $10^{-4}$ annealed every $5$k episodes by a multiplicative factor of $.5$. We set the maximum gradient norm of an update to be $50$. We use an entropy regularization constant of $.001$ for the A2C updates. Even though we have a $.1$ prior over maps, we initialize the belief map with all $.5$ to guarantee that the entropy will decrease. The features are also linearly scaled to be in the range $[-1, 1]$. For all other implementation details, the reader is referred to the source code.

\subsection{Results}

\begin{figure}[h]
    \centering
    \label{fig:learningcurve}
    \caption{Training curves for different network architectures.}
    \includegraphics[scale=.3]{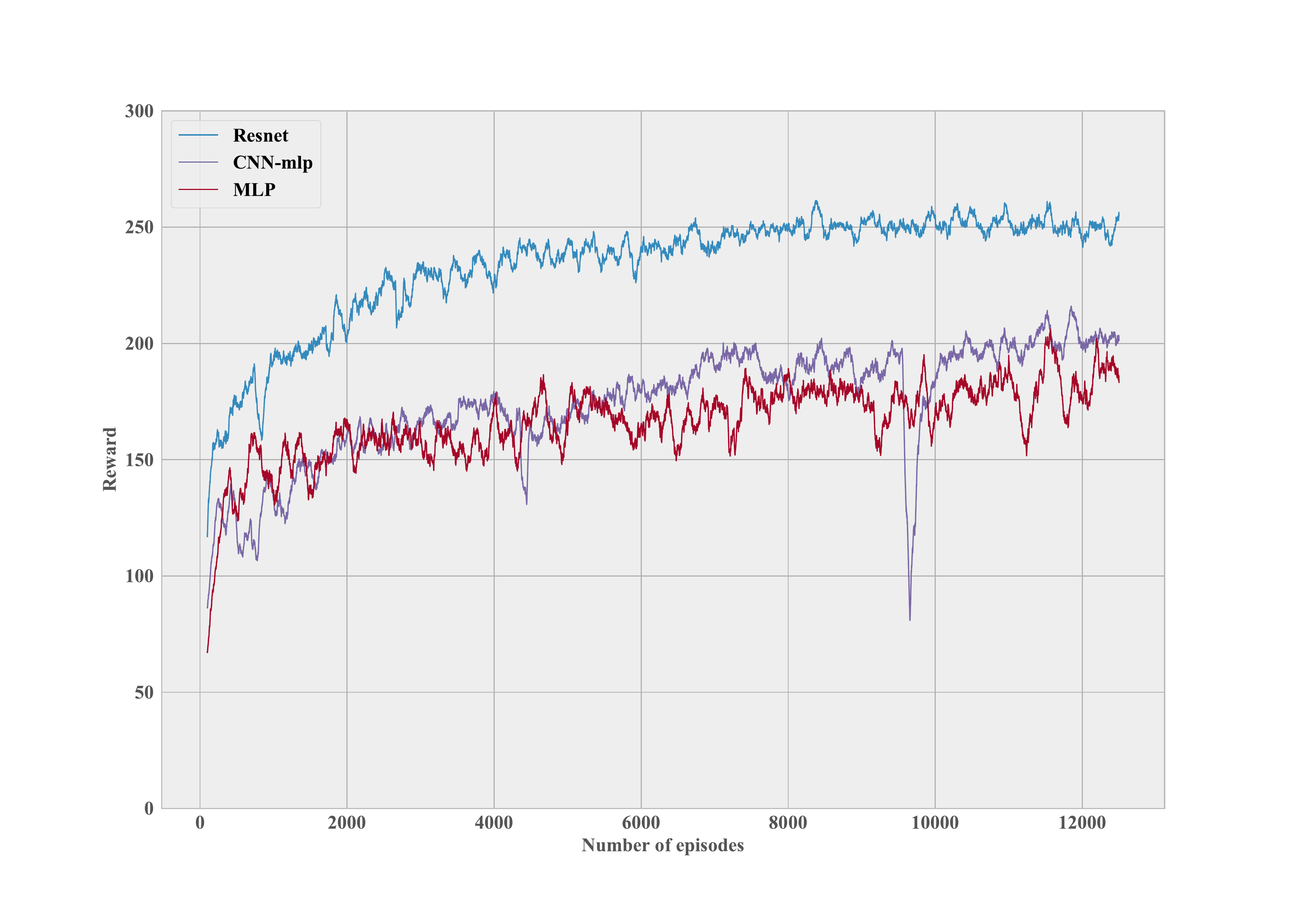}
\end{figure}

We first compare the training characteristics of the three network architectures. The training curves for the three architectures can be found in Figure~\ref{fig:learningcurve}, which contains plots of the episode reward over the number of training episodes for one random seed. The episode rewards have been smoothed with a Gaussian kernel for visualization purposes. The ResNet architecture at first performed the best, which is expected given its superior expressiveness and extensive success in computer vision. However, the training procedure diverged and in fact in the long run the CNN-MLP architecture resulted in a higher average episode reward.

After the training procedure, we evaluate all of the networks, the myopic exploration strategy and a random exploration strategy on $1000$ random, unseen maps. We present the average full episode rewards in Table~\ref{tab:results}. Ultimately, the ResNet architecture outperforms the hand-crafted myopic exploration strategy. This leads us to believe that this training methodology could scale to more complex inverse sensor models where designing myopic exploration strategies is infeasible.

\begin{table}[H]
  \caption{Episode Rewards averaged over 1000 out-of-sample episodes.}
  \label{tab:results}
  \centering
  \begin{tabular}{ll}
    \toprule
    Approach & Performance \\
    \midrule
    ResNet   & $\mathbf{254.87} \ \rpm \ 47.26$  \\
    CNN-MLP  & $201.20 \ \rpm \ 47.25$ \\
    MLP      & $184.61 \ \rpm \ 75.75$ \\
    Myopic   & $251.07 \ \rpm \ 29.2$ \\
    Random   & $92.19 \ \rpm \ 23.84$ \\
    \bottomrule
  \end{tabular}
\end{table}

\section{Conclusion}

In this work, we introduce an approach to train agents that perform efficient robotic mapping using reinforcement learning. This work is preliminary in that it only evaluated the approach on a scenario with a simple inverse sensor model. One could imagine switching the sensor model to be, say, a sonar sensor that depends on the robot's orientation. Then rotating the robot could become part of the action space. Since A2C can be used with continuous action spaces as long as there is an analytical log-probability of actions, this method can be generalized to robotic exploration problems that have continuous action spaces. On the contrary, it is not clear how to generalize information gain-based methods or frontier-based methods to continuous action spaces.

In future work, we hope to apply this approach to problem setups with multiple agents. This is a challenging, unsolved problem, for which we believe our approach could lead to tangible benefits. We would also like to integrate it with recent work on Active Neural Localization \cite{nal}, leading to a fully trainable SLAM system, the holy grail of autonomous navigation robots.

\section*{Acknowledgments}

This material is based upon work supported by the National Science Foundation under Grant No. 2017245257.

\bibliography{mybib}{}
\bibliographystyle{unsrt}

\appendix

\end{document}